\documentclass{bmvc2k}
\usepackage{multirow}
\usepackage{booktabs} 
\usepackage[ruled,vlined]{algorithm2e}

\title{End-to-End Data Visualization by Metric Learning and Coordinate Transformation}

\addauthor{Lilei Zheng}{zhengll@i2r.a-star.edu.sg}{1}
\addauthor{Ying Zhang}{zhangy@i2r.a-star.edu.sg}{2}
\addauthor{Stefan Duffner}{stefan.duffner@liris.cnrs.fr}{1}
\addauthor{Khalid Idrissi}{khalid.idrissi@liris.cnrs.fr}{1}
\addauthor{Christophe Garcia}{christophe.garcia@liris.cnrs.fr}{1}
\addauthor{Atilla Baskurt}{atilla.baskurt@liris.cnrs.fr}{1}

\addinstitution{
Universit\'{e} de Lyon,\\
CNRS, INSA-Lyon, LIRIS,\\
UMR5205, F-69621, France}
\addinstitution{
Cyber Security \& Intelligence,\\
Institute for Infocomm Research, \\
A$^{\star}$STAR, Singapore
}

\runninghead{Zheng \etal}{End-to-End Data Visualization}

\def\etal{\emph{et al}\bmvaOneDot}

\begin{document}

\maketitle

\begin{abstract}
This paper presents a deep nonlinear metric learning framework for data visualization on an image dataset. We propose the Triangular Similarity and prove its equivalence to the Cosine Similarity in measuring a data pair. Based on this novel similarity, a geometrically motivated loss function -- the triangular loss -- is then developed for optimizing a metric learning system comprising two identical CNNs. It is shown that this deep nonlinear system can be efficiently trained by a hybrid algorithm based on the conventional backpropagation algorithm. More interestingly, benefiting from classical manifold learning theories, the proposed system offers two different views to visualize the outputs, the second of which provides better classification results than the state-of-the-art methods in the visualizable spaces.
\end{abstract}

\section{Introduction}
End-to-end learning trains neural networks directly on raw collected data such as speech or image to
achieve effective feature learning for classification or other tasks~\cite{graves2014towards,krizhevsky2012imagenet}. Thus it has recently received much attention because it requires no handcrafted feature extraction algorithms but can directly predict outputs from low-level inputs. For example, for speech recognition, Recurrent Neural Networks (RNN) learn to map from acoustic signals to phonetic sequences~\cite{graves2014towards}; for image classification, Convolutional Neural Networks (CNN) are originally designed to receive an image as input and can classify raw pixels into high-level concepts such as object categories~\cite{krizhevsky2012imagenet}.

Traditional supervised neural networks learn on labeled data samples and
require handcrafted target vectors to conduct error backpropagation, so a CNN classifier is only available for restricted dimensionality reduction because the size of the output layer (i.e. the target dimension) is fixed to the number of classes in a classification problem. This restriction can be relaxed by putting neural networks in a \emph{siamese architecture}~\cite{bromley1993signature} which consists of two identical sub-systems sharing the same set of parameters. This technique, namely Siamese Neural Networks (SNN), performs learning from relationship between data pairs instead of directly on labeled data samples. The SNN approach is categorized to Metric Learning (ML)~\cite{bellet2013survey}, i.e. methods that automatically learn a
metric from a set of data pairs. Therefore, this work proposes a novel metric learning method for end-to-end learning on images.

A good Metric Learning system should be a collaborative product of designing a mapping architecture and formulating a metric-based cost function~\cite{bellet2013survey,chopra2005learning}.
For a mapping function, the depth of the architecture can be shallow or deep, namely, shallow linear transformations~\cite{bellet2013survey}, multi-layer neural networks~\cite{rumelhart1985learning,chen2011extracting,yih2011learning,berlemont2015siamese} or deep neural networks~\cite{bromley1993signature,chopra2005learning,baldi1993neural,hadsell2006dimensionality,mobahi2009deep}; for a cost function, it may be based on a distance metric~\cite{xing2002distance,shalev2004online,weinberger2009distance,goldberger2004neighbourhood,globerson2005metric,davis2007information,guillaumin2009you}, a similarity metric~\cite{qamar2008similarity,qamar2009online,chechik2010large,nguyen2011cosine} or a hybrid metric concerning both distance and similarity~\cite{bar2005learning,cao2013similarity}.



As a commonly used metric for measuring the similarity of two feature vectors, the Cosine Similarity and its variant the bilinear similarity have been found particularly useful for text retrieval~\cite{qamar2008similarity,qamar2009online,chechik2010large,chechik2009online} and face verification~\cite{nguyen2011cosine,cao2013similarity}.
This paper contributes to the study of similarity metric learning and proposes an alternative metric which is equivalent to the Cosine Similarity in measuring a data pair but has a nicer geometrical interpretation of learning the similarity. This novel metric is naturally related to the well-known \emph{triangle inequality theorem}, so we call it the \emph{Triangular Similarity}. The present definition of Triangular Similarity provides a theoretical explanation of the \emph{triangular loss} function that was first given in~\cite{zheng2015triangular}. Moreover, by examining the gradient of this cost function,
we discover its relationship to the Mean Square Error (MSE) function of the traditional neural networks. Therefore typical optimization techniques of training neural networks can be directly applied to optimize our metric learning system.

We integrate the triangular loss function with two CNNs in a siamese architecture, to relax the constraint on the output dimension and make flexible dimensionality reduction to the input data. We name this method as CNN-based Triangular Similarity Metric Learning (TSML-CNN). The method is tested on mapping raw handwritten digit images of the MNIST dataset into visualizable spaces, i.e. target spaces of dimension no larger than 3. This particular kind of dimensionality reduction is so called \emph{data visualization}. The contributions of this paper are the following.
\begin{itemize}
\item[1)] a theoretical proof of learning the Triangular Similarity is explained and illustrated. By a soft length normalization term, we deduce the triangular loss function from the simplified  Triangular Similarity. This completes the theory of TSML.
\item[2)] we present TSML-CNN to perform flexible dimensionality reduction. Compared with that of training a traditional CNN, a significant increase of training time is observed for this siamese variant. However, we introduce a hybrid training strategy to accelerate the training speed.
\item[3)] the proposed method realizes end-to-end data visualization \emph{without} losing its superior ability of accurate classification. It achieves competitive performance on class identification in low dimensional spaces.
\item[4)] we introduce a novel perspective for data visualization in manifold learning field by simple coordinate transformation on the neural networks outputs. Thus we can view the data either on the unit sphere (hypersphere) in a target space, or on the unfolded plane (hyperplane) in an even lower space.
\end{itemize}

\section{Triangular Similarity Metric Learning}
\vspace{-5pt}
\subsection{Triangular Similarity}
For any two given vectors $\textbf{a}$ and $\textbf{b}$, the Triangular Similarity between them is measured by:
\vspace{-3pt}
\begin{equation}
tri(\textbf{a},\textbf{b})=\frac{1}{2}\|\frac{\textbf{a}}{\|\textbf{a}\|}
+\frac{\textbf{b}}{\|\textbf{b}\|}\|,
\vspace{-2pt}
\end{equation}
the value of this similarity lies in the range [0, 1]: it results in 1 if and only if the two vectors $\textbf{a}$ and $\textbf{b}$ are towards the same direction, and yields 0 if and only if the directions of the two vectors are exactly opposite. The Triangular Similarity is related to the Cosine Similarity by:
\vspace{-3pt}
\begin{equation}
\begin{split}
tri(\textbf{a},\textbf{b})&=\frac{1}{2}\|\frac{\textbf{a}}{\|\textbf{a}\|}
+\frac{\textbf{b}}{\|\textbf{b}\|}\|
=\frac{1}{2}\sqrt{(\frac{\textbf{a}}{\|\textbf{a}\|})^2
+(\frac{\textbf{b}}{\|\textbf{b}\|})^2+2\frac{\textbf{a}^T\textbf{b}}
{\|\textbf{a}\|\|\textbf{b}\|}} \\
&=\frac{1}{2}\sqrt{2+2cos(\textbf{a},\textbf{b})}
=\sqrt{\frac{1+cos(\textbf{a},\textbf{b})}{2}},
\end{split}
\vspace{-2pt}
\end{equation}
where $cos(\textbf{a},\textbf{b})$ is the standard Cosine Similarity function.
It can be seen that the relationship between $tri(\textbf{a},\textbf{b})$ and $cos(\textbf{a},\textbf{b})$ is simply a \emph{consecutive} and \emph{bijective} function\footnote{In mathematics, a bijective function is a function between the elements of two sets, where every element of one set is paired with exactly one element of the other set, and every element of the other set is paired with exactly one element of the first set.} $f(z)=\sqrt{(1+z)/2}$ in its effective domain $z\geq -1$, indicating their equivalence in measuring a similarity.

\begin{figure}[t]
\setlength{\belowcaptionskip}{-10pt}
\centering
\includegraphics[width=0.4\textwidth]{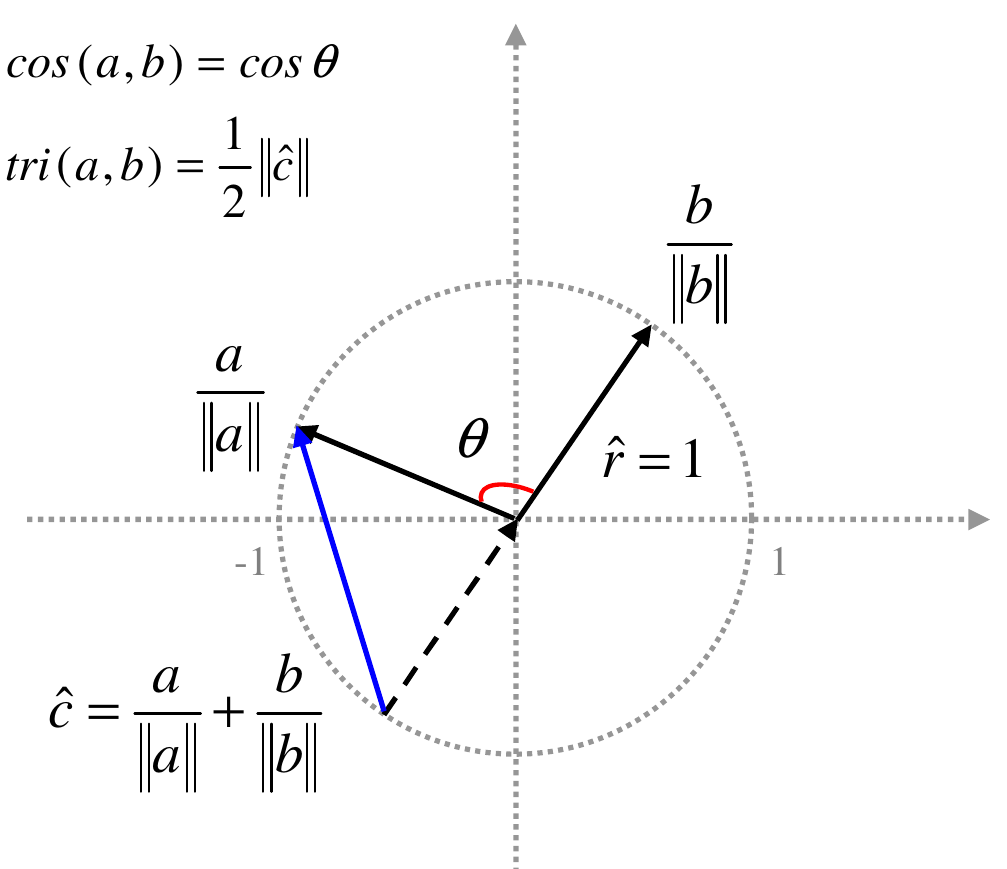}
\vspace{-10pt}
\caption[Illustration of Triangular Similarity]{Equivalence between Triangular Similarity and Cosine Similarity. While the Cosine Similarity simply calculates the \emph{cosine} of the angle $\theta$ between the two vectors, the Triangular Similarity measures the half length of the directed chord (the blue line). The three vectors $\frac{\textbf{a}}{\|\textbf{a}\|}$, $\frac{\textbf{b}}{\|\textbf{b}\|}$ and $\hat{\textbf{c}}$ compose an \emph{isosceles triangle}. The two similarity functions compose a one-to-one correspondence thus the equivalence is confirmed.}
\label{fig3:tri_cos}
\end{figure}

A more intuitive interpretation of the relationship between the two similarities is illustrated in Fig.~\ref{fig3:tri_cos}. In a Cartesian coordinate system,  $\frac{\textbf{a}}{\|\textbf{a}\|}$ and $\frac{\textbf{b}}{\|\textbf{b}\|}$ represent two vectors lying on the unit circle, i.e. a circle of radius one. Their sum determines a directed chord (the blue line) on the circle, denoted by a new vector $\hat{\textbf{c}}$. The Triangular Similarity is simply the half length of the chord, and
the Cosine Similarity calculates the \emph{cosine} of the angle $\theta$ between the two vectors. Note that the three vectors $\frac{\textbf{a}}{\|\textbf{a}\|}$, $\frac{\textbf{b}}{\|\textbf{b}\|}$ and their sum $\hat{\textbf{c}}$ compose an \emph{isosceles triangle}, that is why we call this similarity measurement as the \emph{Triangular Similarity}.

\vspace{-5pt}
\subsection{Triangular Loss Function}
\label{sec3:tiangular}
For both the Cosine Similarity and the Triangular Similarity, the scale of $\|\textbf{a}\|$ and $\|\textbf{b}\|$, i.e. the length of the vectors, should be taken care of. Otherwise, it may raise up numerical instability problems~\cite{higham2002accuracy} especially when $\|\textbf{a}\|$ and $\|\textbf{b}\|$ are too small. This hidden problem can be avoided or relieved by many strategies. For example, adopting regularization terms to prevent $\|\textbf{a}\|$ and $\|\textbf{b}\|$ from degenerating to 0~\cite{berlemont2015siamese}, normalizing the inputs by a whitening transformation~\cite{hyvarinen2000independent,zheng2012acoustic} so that all the input variables have unit variance~\cite{lecun2012efficient}, or simply performing an L2 normalization to let all the vectors have unit length~\cite{cao2013similarity,Simonyan13fisher}. Empirical experiences showed that length regularization enables faster convergence~\cite{lecun2012efficient} and better performance~\cite{cao2013similarity,zheng2012acoustic} to machine learning applications.

We propose a \emph{soft L2 normalization} to constrain the length of the vectors. Instead of performing normalization as a preprocessing step on the inputs,
we constrain the length of the vectors to a constant $r$ by a regularization term:
\vspace{-3pt}
\begin{equation}
\min~~(\|\textbf{a}\|-r)^2+(\|\textbf{b}\|-r)^2.
\vspace{-2pt}
\end{equation}
Minimizing the above function is able to make the values of $\|\textbf{a}\|$ and $\|\textbf{b}\|$ approaching a predefined scalar $r$. But in fact not all the vector lengths can be exactly $r$, so it is regarded as a \emph{soft} length normalization.

With the indicative assumption of $\|\textbf{a}\|\approx \|\textbf{b}\|\approx r$, both the Cosine Similarity and the Triangular Similarity can be simplified. For the Cosine Similarity, it can be rewritten as:
$cos(\textbf{a},\textbf{b})=\frac{\textbf{a}^T\textbf{b}}
{\|\textbf{a}\|\|\textbf{b}\|}\approx (\frac{1}{r})^2\textbf{a}^T\textbf{b}$,
where $\textbf{a}^T\textbf{b}$ is the \emph{bilinear similarity}. This approximation explains that when vectors have approximate lengths, the bilinear similarity is equally effective but simpler than the Cosine Similarity in practical applications~\cite{chechik2010large,deng2011hierarchical,cao2013similarity}.

\begin{figure}[t]
\setlength{\belowcaptionskip}{-10pt}
\centering
\begin{tabular}{cc}
\includegraphics[width=0.48\textwidth]{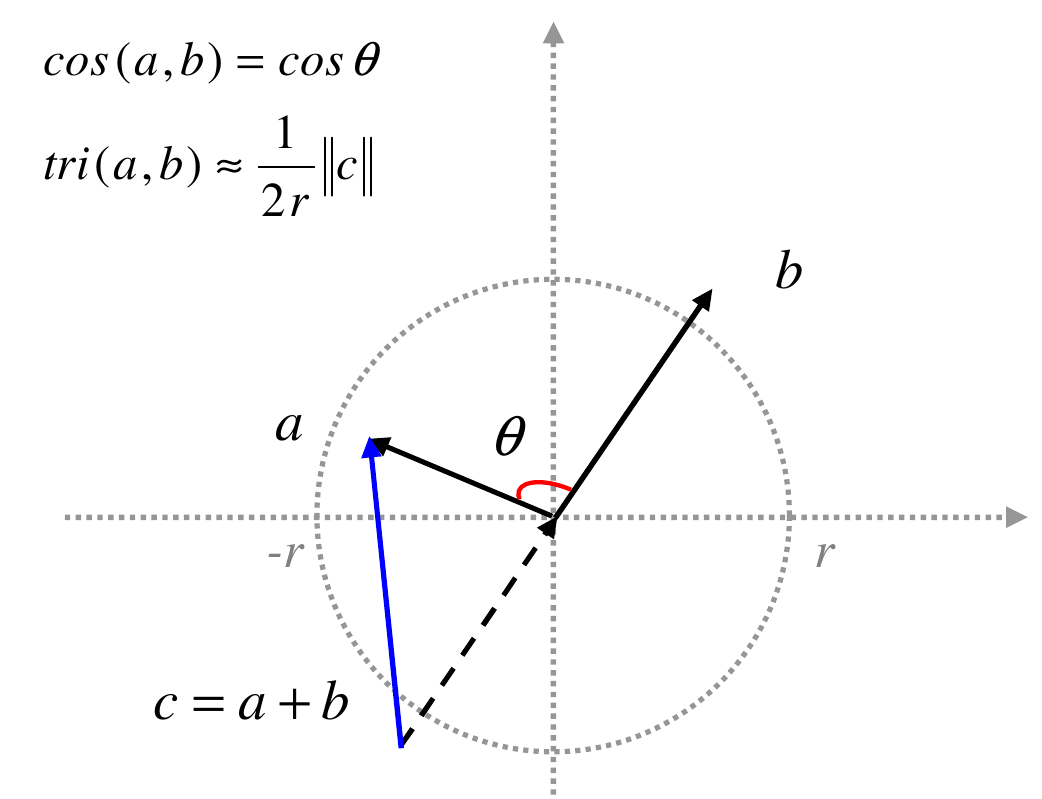}  &
\includegraphics[width=0.48\textwidth]{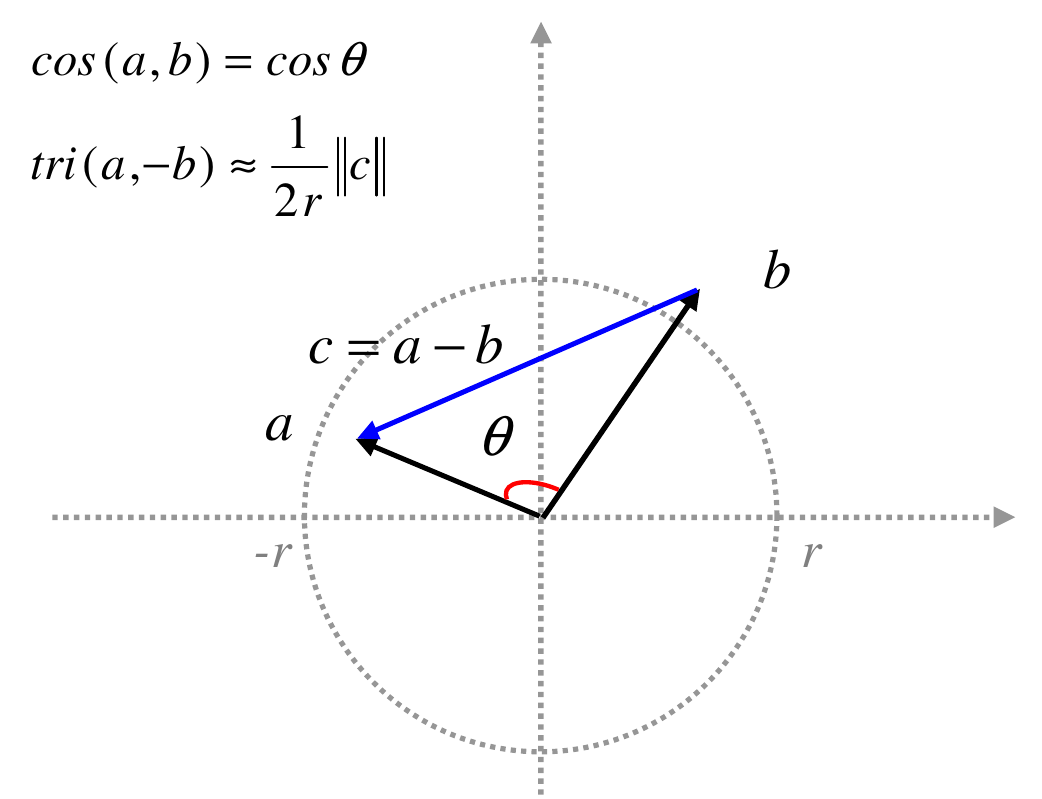} \\
(a) Positive Triangle & (b) Negative Triangle \\
\end{tabular}
\vspace{-5pt}
\caption[Illustration of simplified Triangular Similarity]{The simplified Triangular Similarity concerns normal triangles lying around a circle of radius $r$. A pair of vectors $\textbf{a}$ and $\textbf{b}$ determines two triangles: the positive triangle (left) illustrates the similarity between $\textbf{a}$ and $\textbf{b}$; the negative triangle illustrates the similarity between $\textbf{a}$ and $-\textbf{b}$.}
\label{fig3:two_triangles}
\end{figure}

Similarly, we obtain a simplified version of Triangular Similarity as
$tri(\textbf{a},\textbf{b})=\frac{1}{2}\|\frac{\textbf{a}}{\|\textbf{a}\|}
+\frac{\textbf{b}}{\|\textbf{b}\|}\|\approx \frac{1}{2r}\|\textbf{a}+
\textbf{b}\|$.
Instead of the isosceles triangle in Fig.~\ref{fig3:tri_cos}, this simplified Triangular Similarity concerns a normal triangle lying around a circle of radius $r$ (see Fig.~\ref{fig3:two_triangles} (a)). This triangle is determined by the two vectors $\textbf{a}$ and $\textbf{b}$, completed with their sum $\textbf{c}=\textbf{a}+\textbf{b}$ as the third side. Additionally, the two vectors $\textbf{a}$ and $\textbf{b}$ determine another triangle where the third side is their difference, i.e. $\textbf{c}=\textbf{a}-\textbf{b}$ (Fig.~\ref{fig3:two_triangles} (b)). With respect to the operations of sum or subtraction, we name the two triangles as the positive triangle and the negative triangle, respectively.

Like all the similarity metric learning methods, the objective of learning a Triangular Similarity metric is to increase the similarity between a similar pair and to decrease the similarity between a dissimilar pair. The geometrical interpretation of the objective is the following:
\vspace{-5pt}
\begin{itemize}
\item When $\textbf{a}$ and $\textbf{b}$ are labeled as being similar, increasing the pairwise similarity means minimizing the inter-vector angle $\theta$, which can be realized by maximizing the length of the third side $\textbf{c}$ in the positive triangle (Fig.~\ref{fig3:two_triangles} (a)).
\item When $\textbf{a}$ and $\textbf{b}$ are a dissimilar pair, we need to separte the two vectors with a larger angle $\theta$. This can be also achieved by maximizing the length of the third side $\textbf{c}$ in the negative triangle (Fig.~\ref{fig3:two_triangles} (b)).
\end{itemize}

\vspace{-10pt}
\subsection*{The cost function}
For a set of training data, let $s_i=1$ (respectively -1) denote a pair of vectors $\textbf{a}_i$ and $\textbf{b}_i$ being similar (respectively dissimilar). With the soft length normalization factors $(\|\textbf{a}_i\|-r)^2$, $(\|\textbf{b}_i\|-r)^2$, and the lengths of the three sides, $\|\textbf{a}_i\|$, $\|\textbf{b}_i\|$, $\|\textbf{c}_i\|$, the triangular loss of this pair is defined as:
\vspace{-3pt}
\begin{equation}
J_i=\frac{1}{2}[\boxed{(\|\textbf{a}_i\|-r)^2+(\|\textbf{b}_i\|-r)^2}]+
r(\boxed{\|\textbf{a}_i\|+\|\textbf{b}_i\|-\|\textbf{c}_i\|}),
\label{eq3:tri_twoparts}
\vspace{-2pt}
\end{equation}
where $r$ is a constant constraint for the vector length; $\textbf{c}_i=\textbf{a}_i+s_i\textbf{b}_i$, representing the simplified Triangular Similarity in a positive triangle or a negative triangle. It is interesting to find that the second part of this equation obeys the \emph{triangle inequality theorem}.
Moreover, the coefficients of the two parts are set to $\frac{1}{2}$ and $r$, in order to further simplify the formulation of Equation~(\ref{eq3:tri_twoparts}) as~\cite{zheng2015triangular}:
\vspace{-3pt}
\begin{equation}
J_i=\frac{1}{2}\|\textbf{a}_i\|^2+\frac{1}{2}\|\textbf{b}_i\|^2-r\|\textbf{c}_i\|+r^2.
\label{eq3:tri_general}
\vspace{-2pt}
\end{equation}
%

\vspace{-10pt}
\subsection*{The gradient function}
In Metric Learning systems, the vectors $\textbf{a}_i$ and $\textbf{b}_i$ are outputs of a certain mapping function $f(\cdot)$ parameterized by a set of parameters, e.g. $\textbf{a}_i=f(\textbf{x}_i,\textbf{W})$ where $\textbf{x}_i$ is an input vector. With respect to the parameter set $\textbf{W}$, the derivative of a vector norm is known as $\frac{\partial{\|\textbf{a}\|}}{\partial{\textbf{W}}}
=\frac{\partial{\|\textbf{a}\|}}{\partial{\textbf{a}}}
\frac{\partial{\textbf{a}}}{\partial{\textbf{W}}}
=\frac{\textbf{a}}{\|\textbf{a}\|}\frac{\partial{\textbf{a}}}{\partial{\textbf{W}}}$. Thus the derivative of the triangular loss (Equation~(\ref{eq3:tri_general})) is:
\vspace{-3pt}
\begin{equation}
\frac{\partial{J_i}}{\partial{\textbf{W}}}
=(\textbf{a}_i-r\frac{\textbf{c}_i}{\|\textbf{c}_i\|})
\frac{\partial{\textbf{a}_i}}{\partial{\textbf{W}}}+
(\textbf{b}_i- r\frac{s_i\textbf{c}_i}{\|\textbf{c}_i\|})
\frac{\partial{\textbf{b}_i}}{\partial{\textbf{W}}}.
\label{eq3:tri_gradient}
\vspace{-2pt}
\end{equation}
The partial derivatives $\frac{\partial{\textbf{a}_i}}{\partial{\textbf{W}}}$ and $\frac{\partial{\textbf{b}_i}}{\partial{\textbf{W}}}$ are determined by the specific mapping function $f(\cdot)$, so the ideal minimal cost can be obtained at the zero gradient when
$\textbf{a}_i=r\frac{\textbf{c}_i}{\|\textbf{c}_i\|}$ and
$\textbf{b}_i=r\frac{s_i\textbf{c}_i}{\|\textbf{c}_i\|}$. In other words, the
gradient function has $r\frac{\textbf{c}_i}{\|\textbf{c}_i\|}$ and $r\frac{s_i\textbf{c}_i}{\|\textbf{c}_i\|}$ as targets for $\textbf{a}_i$ and
$\textbf{b}_i$, respectively.

The gradient function supports our predefined objective of learning the similarity, i.e., although the Triangular Similarity in the cost function is only a simplified version as we have assumed the vectors having approximate lengths, we can still achieve that: (1) for two similar vectors, the gradient defines an identical target between them, i.e. $\textbf{a}_i=\textbf{b}_i=r\frac{\textbf{c}_i}{\|\textbf{c}_i\|}$; (2) for two dissimilar vectors, the gradient projects them to opposite directions, i.e. $\textbf{a}_i=r\frac{\textbf{c}_i}{\|\textbf{c}_i\|}$ and $\textbf{b}_i=-r\frac{\textbf{c}_i}{\|\textbf{c}_i\|}$.
%

\vspace{-5pt}
\subsection{Relation to Traditional Neural Networks}
%
For classification problems, it is popular to use the Mean Squared Error (MSE) cost function
in traditional neural networks, either an MLP~\cite{rumelhart1985learning} or a CNN~\cite{lecun1998gradient}.
It simply measures the difference between a computed output of a network and its desired target.
Formally, for a given training sample $\textbf{x}_i$ and its predefined target $\textbf{g}_i$, we first compute its output by the mapping function, i.e. $\textbf{a}_i=f(\textbf{x}_i,\textbf{W})$. The error for this training sample is the squared Euclidean distance between $\textbf{a}_i$ and $\textbf{g}_i$. The cost function and its gradient are:
\vspace{-3pt}
\begin{equation}
J_i=\frac{1}{2}(\textbf{a}_i-\textbf{g}_i)^2;~~~~
\frac{\partial{J_i}}{\partial{\textbf{W}}}
=(\textbf{a}_i-\textbf{g}_i)
\frac{\partial{\textbf{a}_i}}{\partial{\textbf{W}}}.
\vspace{-2pt}
\end{equation}

One can see that the gradient function of the triangular loss (Equation~(\ref{eq3:tri_gradient})) is exactly \emph{a double copy} of the MSE gradient. The difference is that (1) the single output $\textbf{a}_i$ in traditional neural networks is now paired with a partner $\textbf{b}_i$ to learn the pairwise relationship between data; (2) the handcrafted target $\textbf{g}_i$ is replaced by temporal targets $r\frac{\textbf{c}_i}{\|\textbf{c}_i\|}$ and $r\frac{s_i\textbf{c}_i}{\|\textbf{c}_i\|}$ which are automatically specified by the two vectors $\textbf{a}_i$ and $\textbf{b}_i$ themselves. This is indeed an advantage that
the dimension of the output vectors is no longer required to be equal to the number of classes, and thus flexible dimensionality reduction is achieved by our metric learning system.
Furthermore, with the similar gradient formulations, typical optimization techniques and practical tricks of training neural networks~\cite{orr2003neural} -- such as Weight Decay, Momentum~\cite{lecun2012efficient}, Dropout~\cite{srivastava2014dropout} -- can be directly applied to optimize our system.

\vspace{-5pt}
\subsection{Comparison to Euclidean Distance}
When we know that the Triangular Similarity is equivalent to the Cosine Similarity in measuring a data pair, one may care about the comparison between the Triangular Similarity and the Euclidean distance. The Euclidean distance is
naturally related to the Cosine Similarity: for the Euclidean distance between two vectors, $\|\textbf{a}-\textbf{b}\|$, we know that
$\|\textbf{a}-\textbf{b}\|^2=(\textbf{a}-\textbf{b})^T(\textbf{a}-\textbf{b})=
\textbf{a}^2+\textbf{b}^2-2\textbf{a}^T\textbf{b}$. If the
vectors are normalized to unit length, i.e. $\textbf{a}^2=\textbf{b}^2=1$, this equation reads that $\|\textbf{a}-\textbf{b}\|^2 = 2-2cos(\textbf{a},\textbf{b})$. That means, since we have controlled the vector length in our triangular loss function, minimizing the pairwise similarity value is approximately equivalent to maximizing the Euclidean distance.

However, whether the vector length is controlled or not, the advantage of the Cosine Similarity
(as well as the Triangular Similarity) over the Euclidean distance is that it does not use the length information of vectors to distinguish two vectors. In other words, only the degree of freedom on the angle is enough to represent and distinguish vectors in different classes.
In our similarity metric learning system,
an operation of length normalization can be therefore adopted to removes length information of output vectors.
This operation is especially interesting and useful for data visualization. Without the degree of freedom on the length, outputs are mapped to
a sphere or hypersphere in the target space. According to the manifold learning theory that a hypersphere with a point removed is homeomorphic with a hyperplane~\cite{mendelson1990introduction}, we can unfold the hypersphere into a lower dimensional space so that we obtain a new perspective to view the mapping result. This unfolding is simply realized by \emph{coordinate transformation}, in a similar way of drawing a world map for the earth.

\section{End-to-end Data Visualization on Large-scale Data}
\label{sec5:large_scale}
\vspace{-5pt}
\subsection{The MNIST Dataset}
The MNIST handwritten digits dataset~\cite{lecun1998gradient}\footnote{http://yann.lecun.com/exdb/mnist/} is a popular benchmark for classification by neural network based algorithms.
There are 70,000 8-bit grayscale images in total, where 60,000 images are separated as the training data and the remaining 10,000 images are used for testing. All the images are of the same size $28\times 28$, capturing a digit from 0 to 9 with various writing styles.

\subsection{TSML-CNN}
\begin{figure}[t]
\setlength{\belowcaptionskip}{-10pt}
\centering
\includegraphics[width=\textwidth]{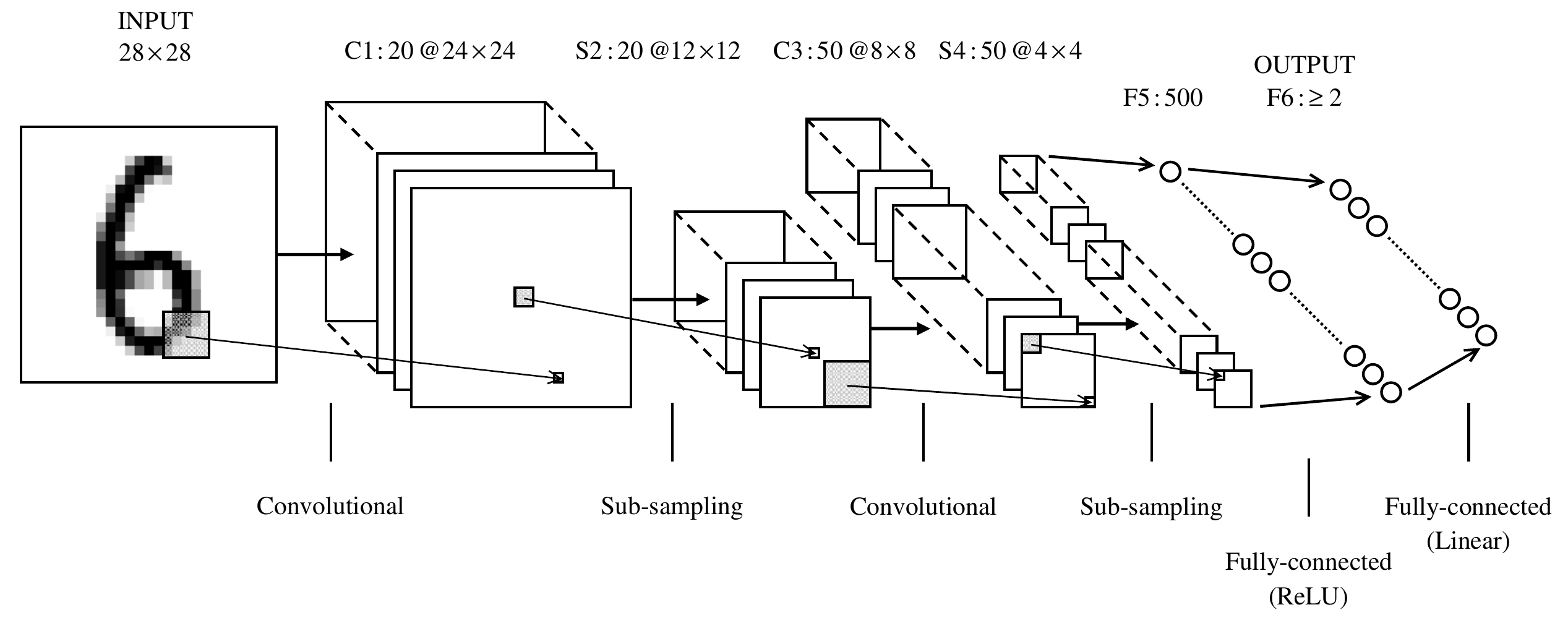}
\vspace{-25pt}
\caption[Diagram of the proposed CNN architecture]{Diagram of our proposed CNN architecture.}
\label{fig5:myCNN}
\end{figure}

CNNs are a specialized kind of neural networks for processing data that have a known \emph{grid-like topology}~\cite{Bengio-et-al-2015-Book}, i.e. 2-dimensional data such as images and speech time-series~\cite{lecun1998gradient,ning2005toward,lawrence1997face,lecun2010convolutional}. In this work, we employ the CNN architecture recommended by~\cite{jia2014caffe}\footnote{http://caffe.berkeleyvision.org/} to process the images of the MNIST dataset and train our TSML system. In particular, this CNN architecture is similar with that in LeNet-5~\cite{lecun1998gradient} but of different hyperparameters. Figure~\ref{fig5:myCNN} illustrates the layers in this CNN.

This CNN is composed of 6 layers, not counting the input layer. There are two convolutional layers at the 1st and 3rd layers, denoted by C1 and C3. Particularly, C1 and C3 are followed by two sub-sampling layers (i.e. pooling layers) S2 and S4.
For all the convolutional layers, the size of local receptive fields is always $5\times5$. And a feature map is learned on one or more pages of units in the previous layer.
For all the sub-sampling layers, the size of pools is always $2\times2$. Especially, we adopt a nonparametric sub-sampling operation, i.e. the max pooling: the four inputs in a pool are compared and the maximum is delivered to the following layer.
The 5th layer F5 is a fully-connected layer having a ReLU activation function~\cite{glorot2011deep}. It contains $500$ hidden units that are connected to every hidden unit in S4.
The 6th layer F6 is also a fully-connected layer but performing a simple linear mapping into the target space. The size of this layer depends on the choice of the target space. For example, if we want to realize a mapping into the 3-dimensional space, we set the size of this layer to 3. We regard this layer as the output layer as we use the element values in this layer as the final vector representation for the input image.

Let a matrix $\textbf{X}$ denote an input image and a vector $\textbf{a}$ represent the output vector, the above CNN accomplishes a deep nonlinear mapping $f(\cdot)$ that $\textbf{a}=f(\textbf{X},\textbf{W})$ where $\textbf{W}$ indicates all the parameters in this CNN. Involving the CNN in our siamese architecture, TSML-CNN simply takes a pair of images $\textbf{X}_i$ and $\textbf{Y}_i$ from a training set and produces two outputs $\textbf{a}_i=f(\textbf{X}_i,\textbf{W})$ and $\textbf{b}_i=f(\textbf{X}_i,\textbf{W})$. TSML-CNN calculates the triangular loss on the two outputs by Equation~(\ref{eq3:tri_general}). However, training a siamese network is much slower than training a traditional single-track network of the same size~\cite{zheng2015siamese}. Especially, in the large training set of the MNIST dataset, the 60,000 training samples indicate 1.8 billion training pairs in total, thus the time consumption of directly training TSML-CNN by data pairs is unaffordable.

\vspace{-5pt}
\subsection{Hybrid Training: TSML-Hybrid}
To provide an efficient dimensionality reduction, we propose a hybrid training algorithm, namely TSML-Hybrid that includes the following four stages:
\vspace{-3pt}
\begin{itemize}
\item[1)] \textit{Tiny-scale training}: select only a few training samples from each class and train TSML-CNN on corresponding similar and dissimilar training pairs. In practice, the similar pairs play an important role in controlling vector lengths, thus we usually select at least 2 training samples from each class in order to generate both similar and dissimilar pairs. On the few training data, TSML-CNN gets convergence to an optimal solution very quickly.
\item[2)] \textit{Transplanting}: set the centers of different classes in the target space as new labels for each class; take the learned parameters in TSML-CNN as initialization for a new CNN. This operation can be considered as transplanting the CNN in TSML-CNN to a new single-track CNN. With this transplantation, the single-track CNN needs no handcrafted target vectors and thus inherits the ability of flexible dimensional reduction from TSML-CNN.
\item[3)] \textit{Large-scale training}: take the new labels as target vectors for each class and train the single-track CNN on all training samples. The MSE function computes cost between output vectors and their desired targets. Compared with the large number of possible data pairs, the number of data samples is much smaller thus efficient convergence can be reached.
\item[4)] \textit{Length normalization}: apply L2 normalization to make all the output vectors having unit length in the target space, because the Cosine Similarity
    (as well as the Triangular Similarity) does not need the length information of vectors to distinguish different classes.
\end{itemize}
Compared to directly training a siamese network such as TSML-CNN, the proposed hybrid algorithm greatly improves the efficiency of training but does not suppress the classification performance. We will provide more results in the experimental section.

\vspace{-5pt}
\section{Results and Analysis}
In this section, we visualize the MNIST digits in the 2/3/4-dimensional spaces by using HSML-Hybrid.
%
%

\begin{figure}[t]
\setlength{\belowcaptionskip}{-10pt}
\centering
\begin{tabular}{cccc}
\includegraphics[width=0.22\textwidth]{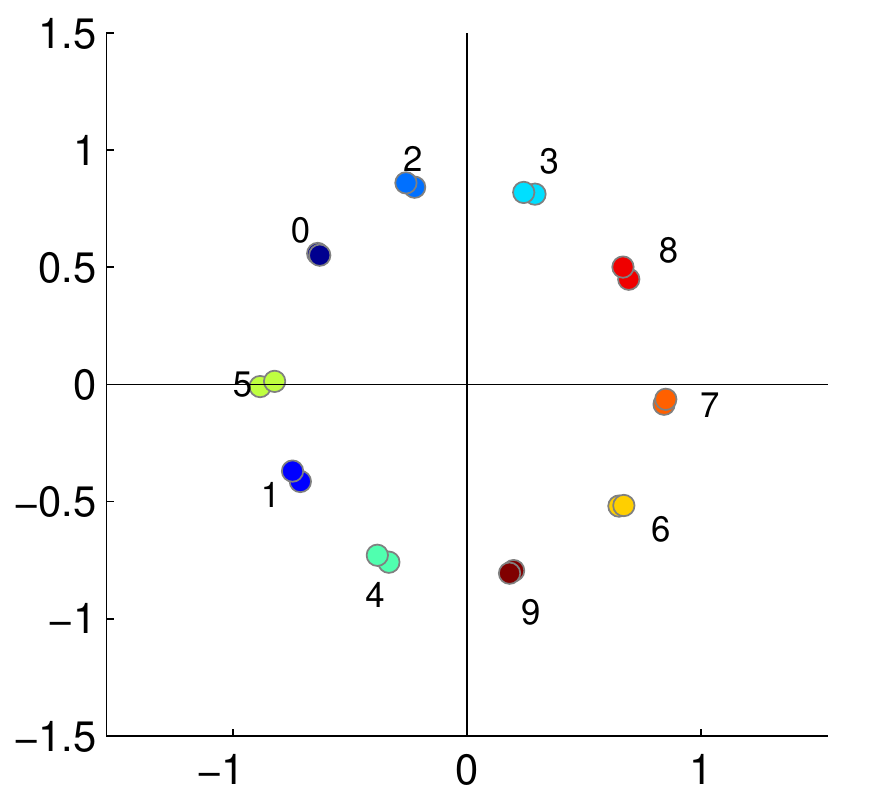} &
\includegraphics[width=0.22\textwidth]{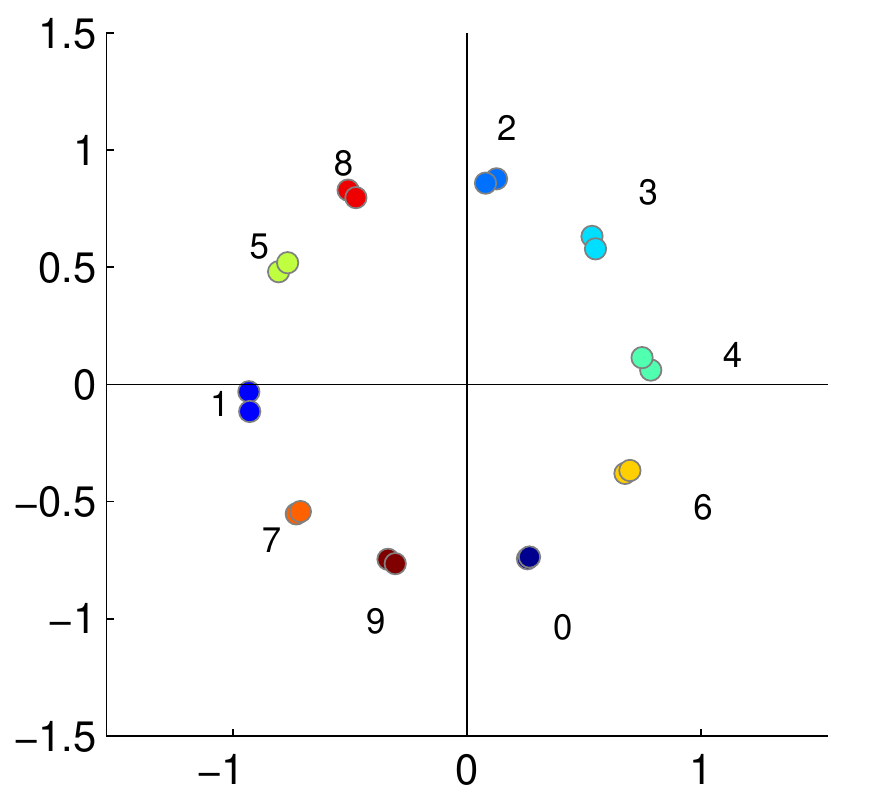} &
\includegraphics[width=0.22\textwidth]{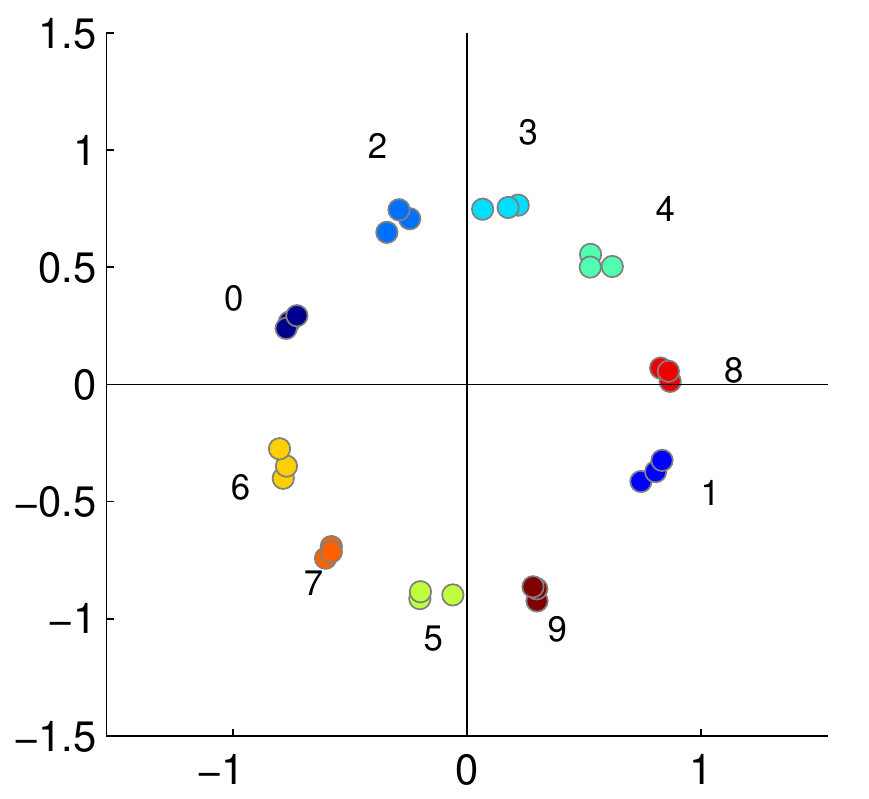} &
\includegraphics[width=0.22\textwidth]{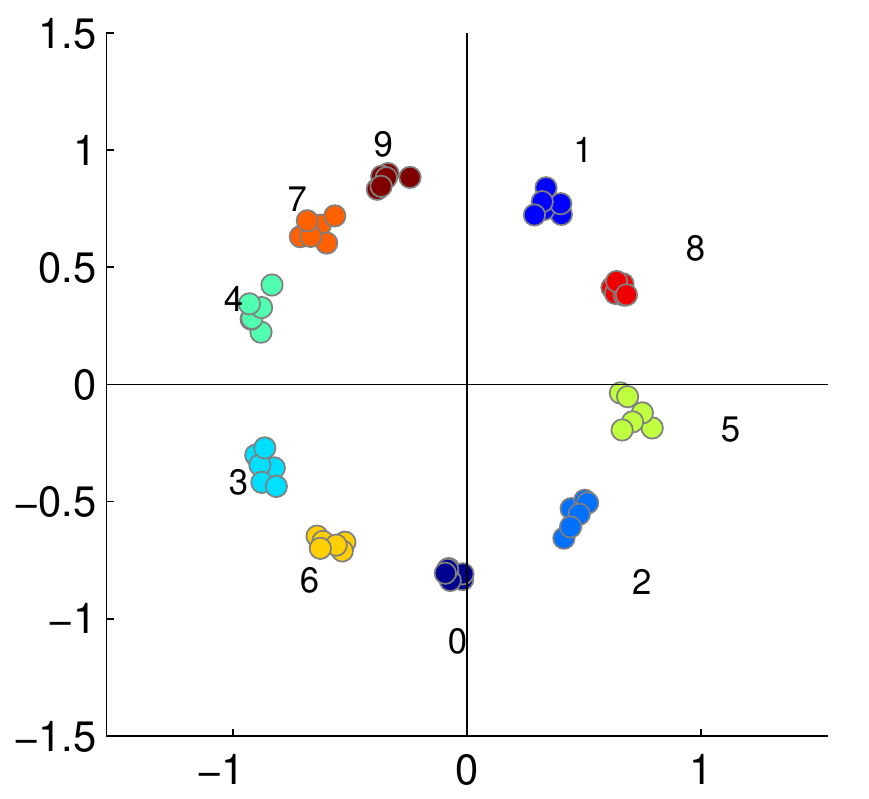} \\
(a) 20 samples & (b) 20 samples & (c) 30 samples & (d) 60 samples
\end{tabular}
\vspace{-10pt}
\caption{Results after tiny-scale training with different data size and initialization.}
\label{fig5:tsmlCNN2naive}
\end{figure}

\begin{figure}[t]
\centering
\begin{tabular}{ccc}
\includegraphics[width=0.3\textwidth]{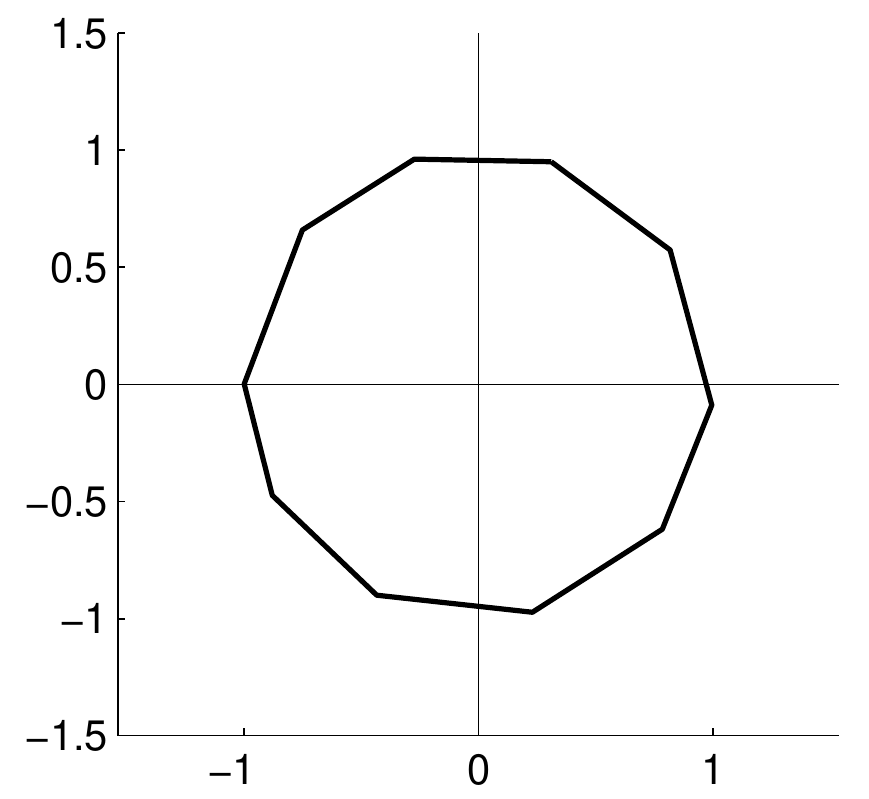} &
\includegraphics[width=0.288\textwidth]{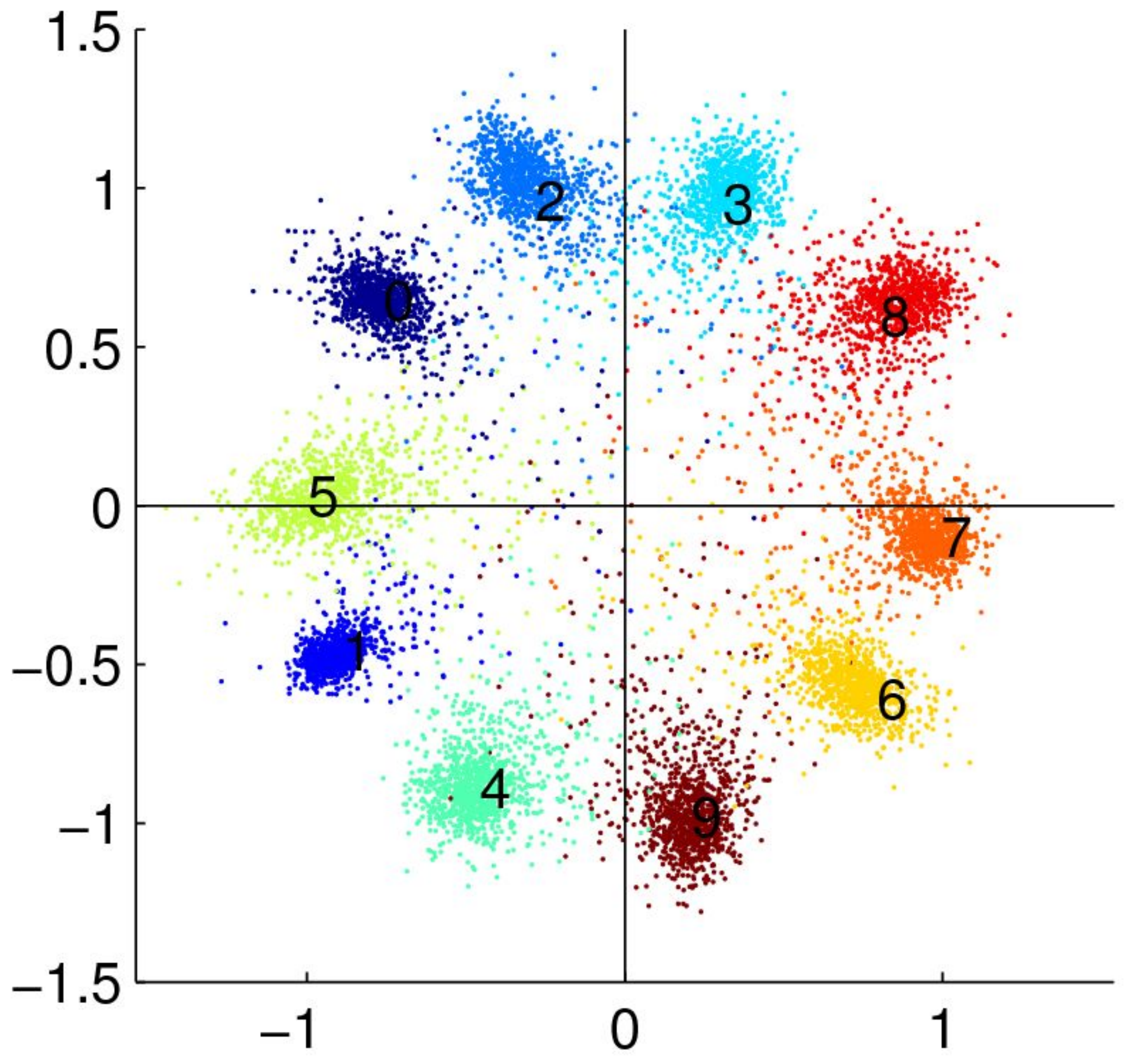} &
\includegraphics[width=0.288\textwidth]{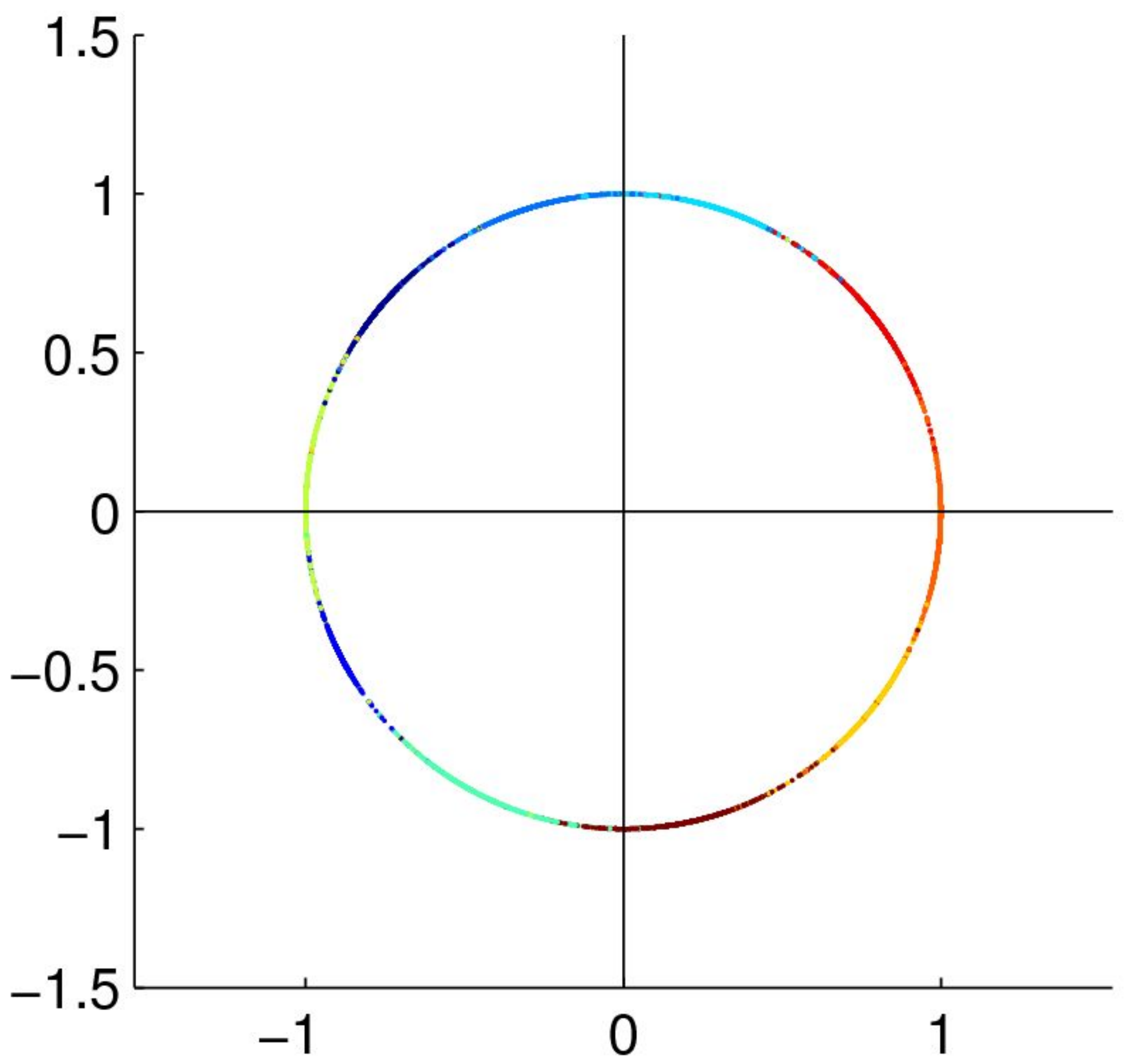} \\
(a) & (b) & (c)
\end{tabular}
\caption{After tiny-scale training, (a) centers of different classes stand as vertexes of a polygon; after large-scale training, (c) the 10,000 test data assemble around each center; by length normalization, (d) the data are further projected onto the unit circle.}
\label{fig5:tsmlCNN2final}

\centering
\includegraphics[width=\textwidth]{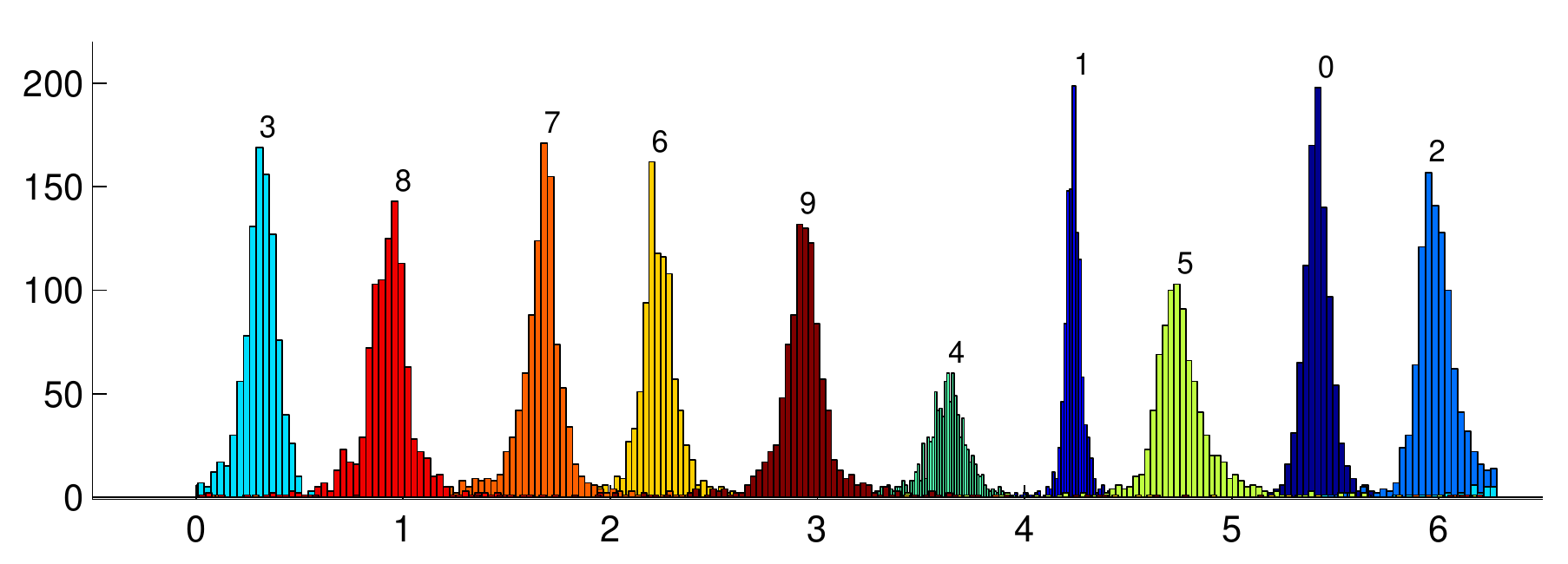} \\
\includegraphics[width=1.08\textwidth, trim=59 10 10 0]{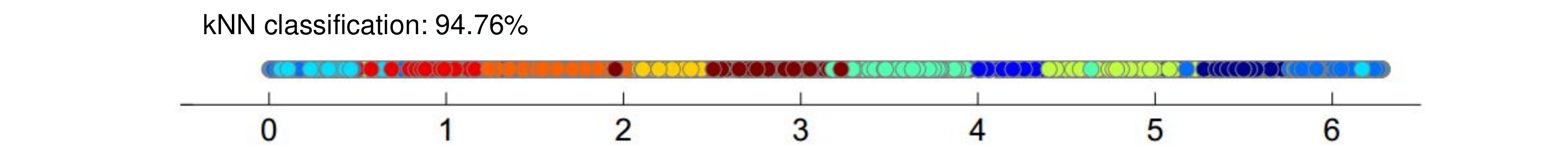}
\caption[An unfolded view of the MNIST test data (2-dimensional)]{The lower picture gives an unfolded view of the MNIST test data in Fig.~\ref{fig5:tsmlCNN2final} (c), i.e. an interval $[0,2\pi]$; the upper picture plots the histograms of the 10 classes.}
\label{fig5:tsmlCNN2unfold}
\end{figure}

\vspace{-5pt}
\subsection{Results of Data Visualization}
\vspace{-5pt}
\subsubsection{Case study 1: 2-dimension}
In \emph{tiny-scale training}, we randomly select 2 training sample from each of the 10 classes in the MNIST dataset. We train TSML-CNN on these data pairs with the output dimension equal to 2. The learned model projects the 20 training samples into the target space as in Fig.~\ref{fig5:tsmlCNN2naive} (a), where numbers beside the points denote the digits they represent. The positions of the digits are determined by the random initialization and tiny-scale training procedure. We observe no evident relation between adjacent digits in our experiments: Fig.~\ref{fig5:tsmlCNN2naive} shows other mapping results with different data size and initialization, though the order of the 10 digits varies, the overall distribution is always approximating a circle.

After the tiny-scale training, the 20 samples evenly distribute along a circle around the origin, connecting the centers of every class results in a 10-sided polygon (Fig.~\ref{fig5:tsmlCNN2final} (a)).
In \emph{large-scale training}, the class centers are set as labels (i.e. target vectors) for their corresponding classes. Minimizing the MSE cost function attracts all the training samples to these centers. The final distribution of the 10,000 test data after this training is shown in Fig.~\ref{fig5:tsmlCNN2final} (b). We can see that data of each class scatter in a local small area in the 2-dimensional space, assembling around the labeled class center. By length normalization, the data vectors are then projected to the unit circle (see Fig.~\ref{fig5:tsmlCNN2final} (c)).

In manifold learning theory, a circle with a point removed is homeomorphic with a real line. In other words, the points on the circle can be represented by the degree of freedom on the angle only. In Figure~\ref{fig5:tsmlCNN2unfold}, we unfold the circle to an interval $[0,2\pi]$ in the lower picture, that means each test image is represented by a single point within this interval. In the upper picture, the histograms of the test data in the 10 classes are shown. We can see that each class has a sharp peak and adjacent classes are divided with clear valleys, i.e. data in the same class have been well concentrated and data in different classes have been well separated. One may notice the shape of different histograms varies, e.g. the peak of the number 4 is low and fatty but that of the number 1 is high and sharp. We found that the shape of the histogram -- the final distribution of the output data -- is automatically determined by the random initialization and local optimal convergence of the present neural networks.

\begin{figure}
\centering
\begin{tabular}{cc}
\includegraphics[width=0.4\textwidth]{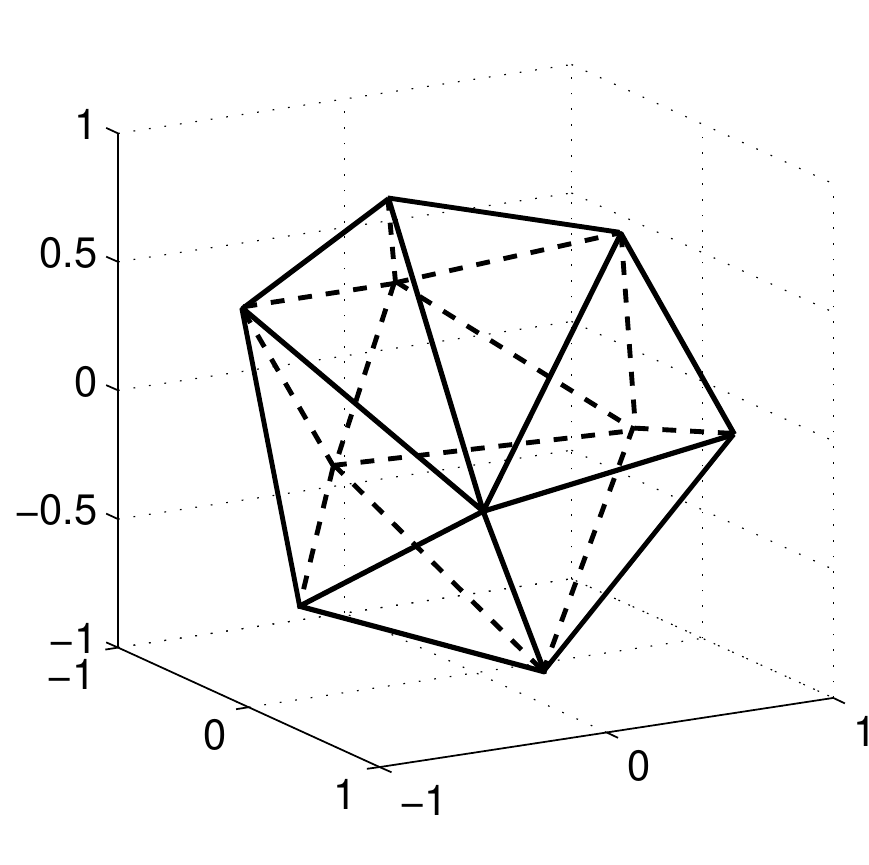} &
\includegraphics[width=0.4\textwidth]{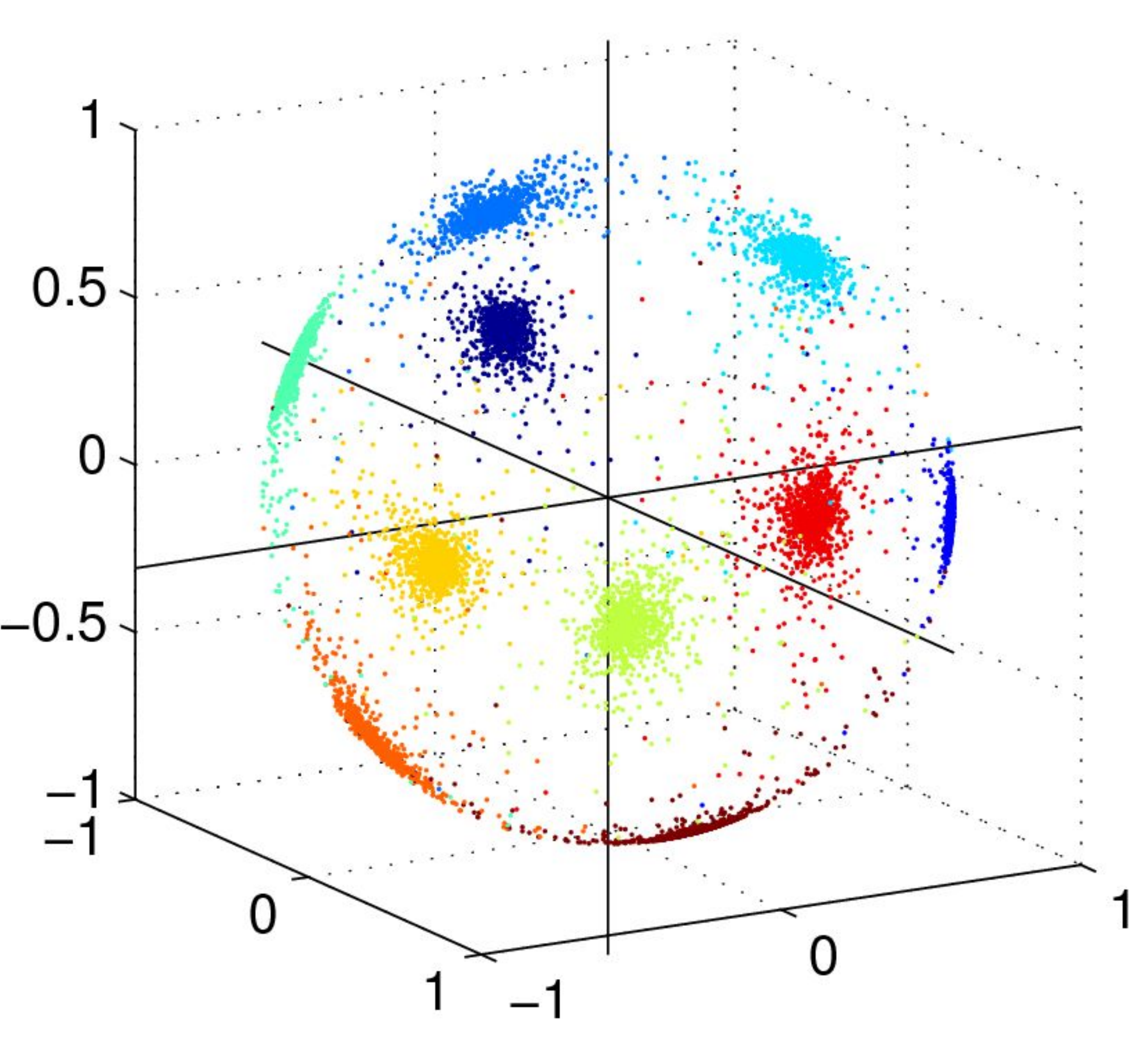}\\
(a) Polyhedron connecting the target centers & (b) Test data after large-scale training \\
\multicolumn{2}{c}{\includegraphics[width=0.8\textwidth]{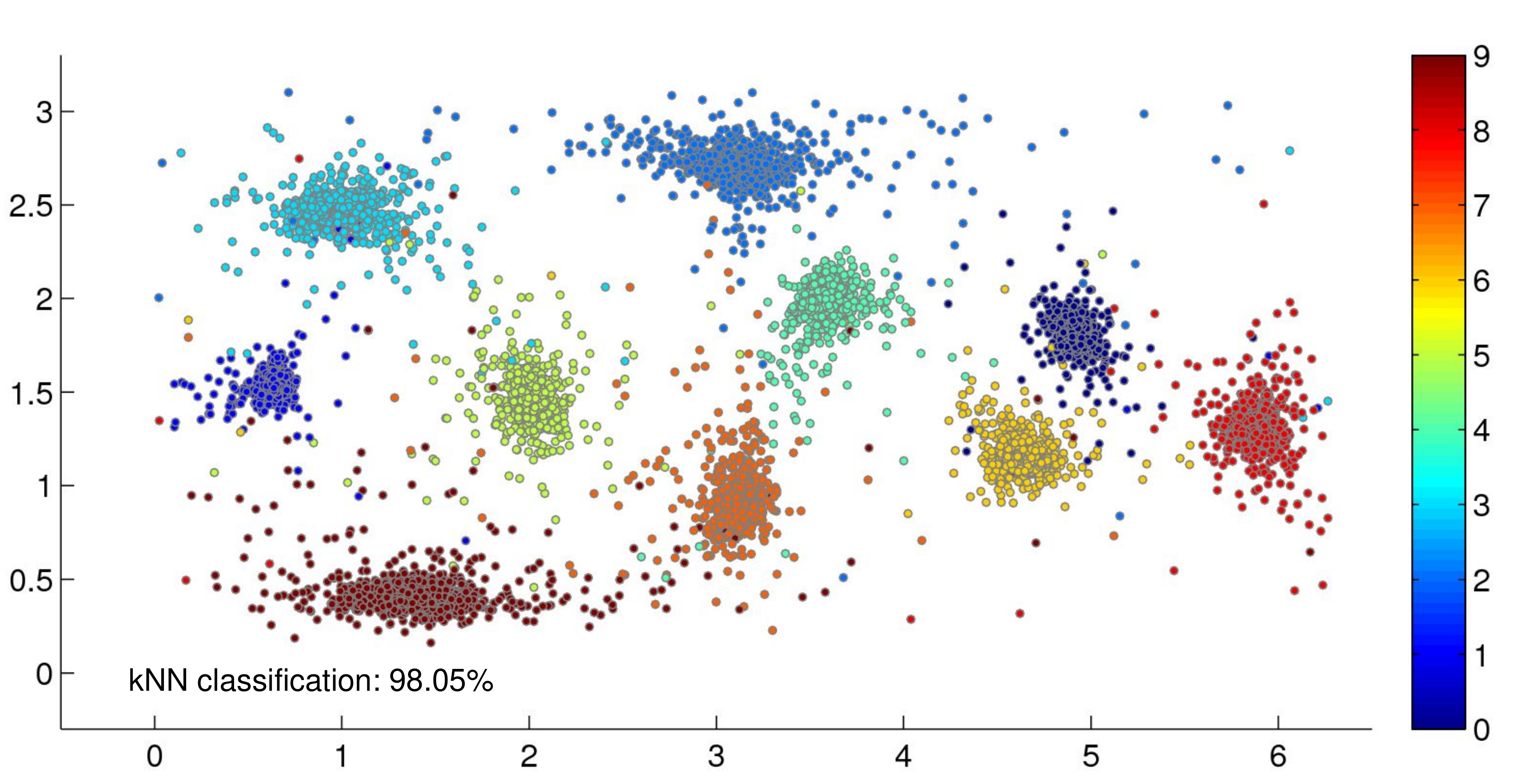}} \\
\multicolumn{2}{c}{(c) Unfolded view of the test data} \\
\end{tabular}
\caption[Mapping results in the 3-dimensional target space.]{Mapping results in the 3-dimensional target space.}
\label{fig5:tsmlCNN3}
\centering
\includegraphics[width=0.8\textwidth]{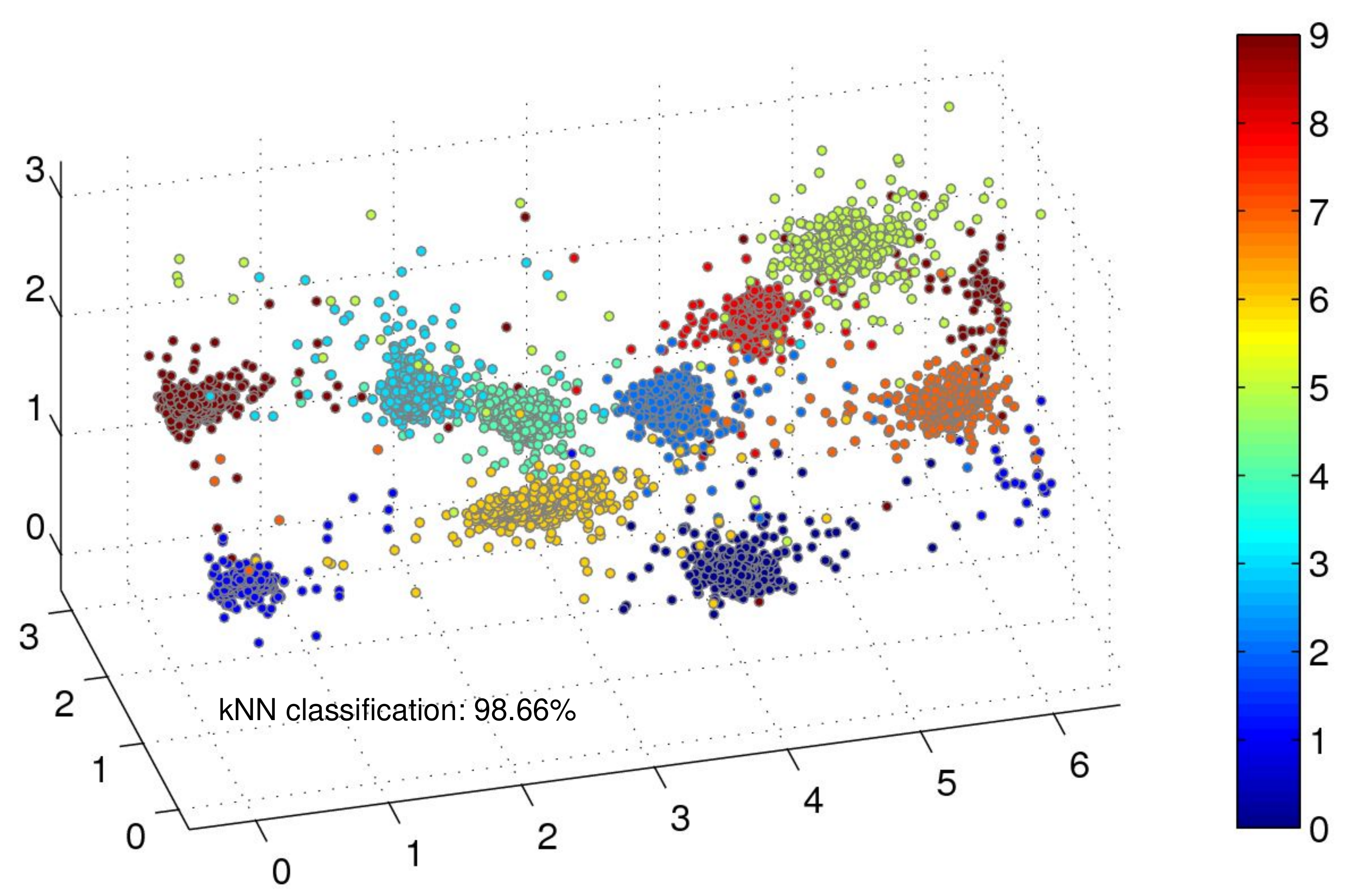}
\caption[An unfolded view of the MNIST test data (4-dimensional)]{The unfolded view of the 4-dimensional MNIST test data in the 3-dimensional space.}
\label{fig5:tsmlCNN4unfold}
\end{figure}

However, recent theoretical and empirical results strongly suggest that the different local optima are not a serious issue in general~\cite{lecun2015deep}. Regardless of initial conditions that generate different output distributions, our non-convex system nearly always reaches local solutions of very similar classification quality.
Specifically, $k$NN classification on either the 2-dimensional data on the circle or the 1-dimensional data within the interval yields an accuracy of $94.76\%$.
In conclusion, with only one mapping by the proposed similarity metric learning, we have obtained two views for visualizing the test data, i.e. Fig.~\ref{fig5:tsmlCNN2final} (c) and Fig.~\ref{fig5:tsmlCNN2unfold}.

\vspace{-10pt}
\subsubsection{Case study 2: 3-dimension}
When we set the output dimension to 3, a different model is learned on the same 20 training samples. Firstly, the well-trained TSML-CNN projects the 20 training samples into the 3-dimensional space, comprising a convex polyhedron with 10 vertexes that will be later used as target vectors for the 10 classes, as shown in Fig.~\ref{fig5:tsmlCNN3} (a).

Similar with the unit circle in the case of 2-dimension, the TSML-Hybrid model produces a unit sphere carrying all the 3-dimensional data (see Fig.~\ref{fig5:tsmlCNN3} (b)). More interestingly, a sphere with a point removed is homeomorphic with a plane. In other words, considering the sphere as the earth, the plane is like the world map. Figure~\ref{fig5:tsmlCNN3} (c) illustrates the unfolded map for the test data. On this map, different classes of data locate as 10 isolated continents. Note that the leftmost part of the map and the rightmost part are actually connected in the 3-dimensional space. The length and width of this map are $2\pi$ and $\pi$, respectively. In addition, $k$NN classification accuracy of the 2-dimensional data is $98.05\%$, much higher than that of the 1-dimensional data ($94.76\%$).

\vspace{-10pt}
\subsubsection{Case study 3: 4-dimension}
In fact, we can not see the 4-dimensional space but only a subspace of it. Fortunately, the proposed method provides such a subspace of the target space, i.e. the unit hypersphere. Unfolding this 4-dimensional hypersphere results in a part of the 3-dimensional space (see Fig.~\ref{fig5:tsmlCNN4unfold}). The length, width and height of this partial space are $2\pi$, $\pi$ and $\pi$, respectively. Like the previous cases, we observe clear boundaries between different classes. One may notice that a few data of the two classes at the leftmost side are mapped to the rightmost side, actually they are together in the 4-dimensional space. In terms of better data visualization, a probable solution can be duplicating the 3-dimensional mapping results to make the visualization locally complete. In terms of classification, since the $k$NN classifier makes prediction on the majority label of $k$-nearest neighbors in the training set, this phenomenon of separated data has little influence on the classification performance. The $k$NN classifier obtains an classification accuracy of $98.66\%$ on the unfolded 3-dimensional data.

\begin{figure}[t]
\setlength{\belowcaptionskip}{-12pt}
\centering
\begin{tabular}{cc}
\includegraphics[width=0.43\textwidth]{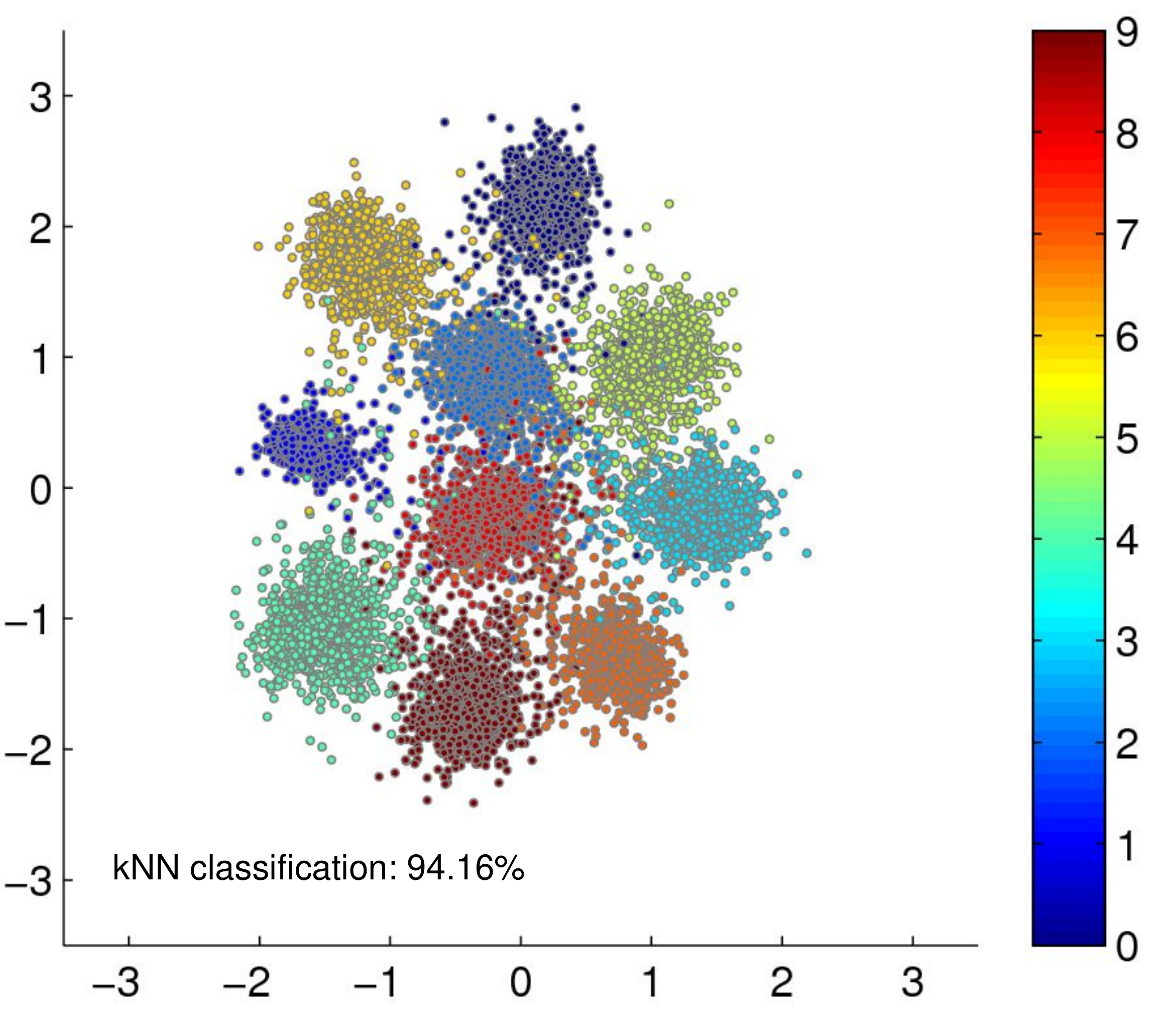} &
\includegraphics[width=0.5\textwidth]{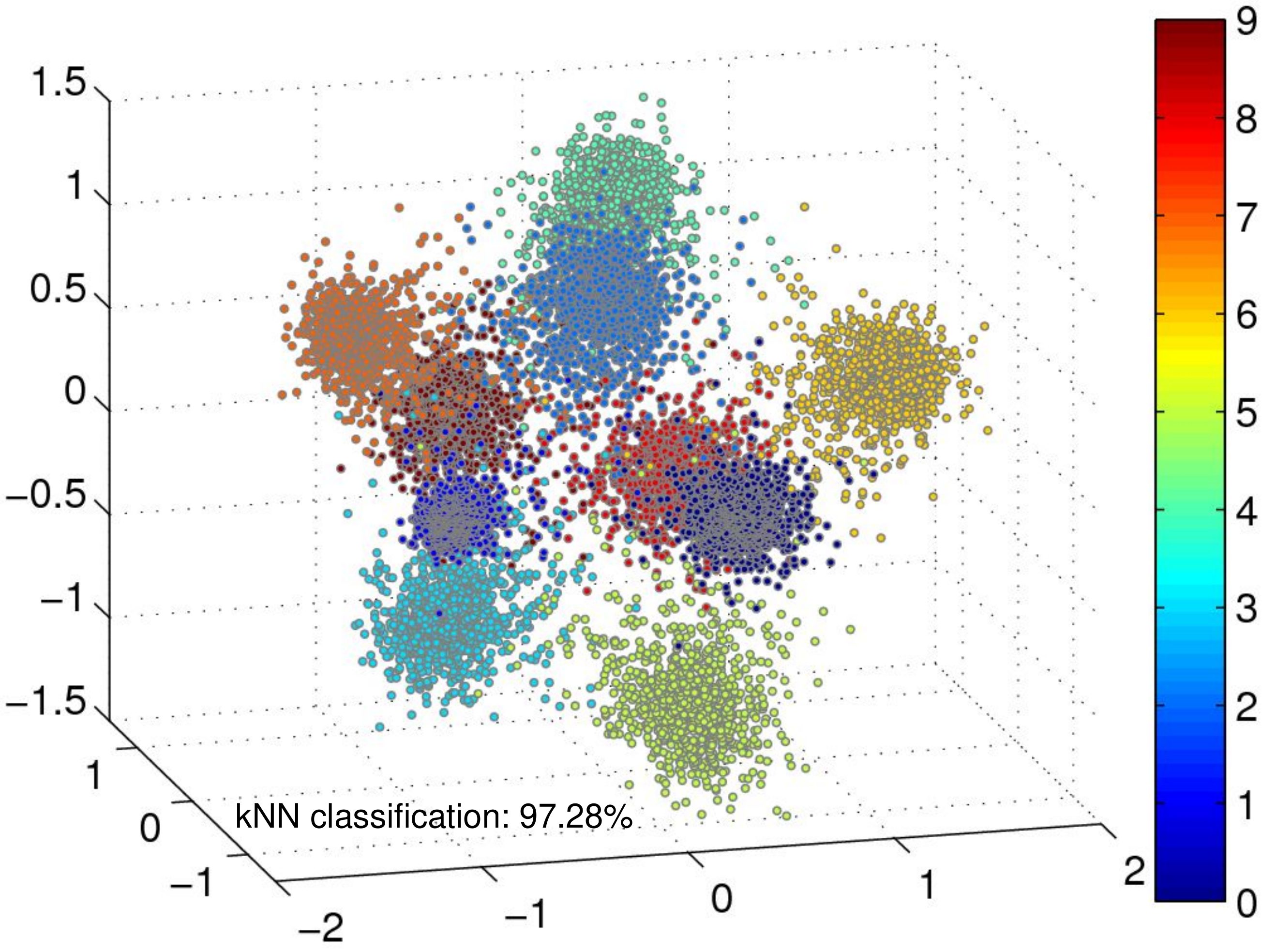}\\
(a) In 2-dimensional space &
(b) In 3-dimensional space \\
\end{tabular}
\vspace{-7pt}
\caption[Mapping results of the MNIST test data by DrLIM]{Mapping results by DrLIM in (a) 2-dimensional and (b) 3-dimensional spaces.}
\label{fig5:drlimView}
\end{figure}

\vspace{-7pt}
\subsection{Comparison to the State-of-the-art}
\vspace{-5pt}
The most related work with ours is Dimensionality Reduction by Learning an Invariant Mapping (DrLIM)~\cite{hadsell2006dimensionality}, which proposed a contrastive loss function to measure the cost of data pairs. Different with our triangular loss function concerning the Triangular Similarity (or the Cosine Similarity), the contrastive loss function was based on the Euclidean distance. The objective of the contrastive loss is to minimize the distance between a similar pair and to separate any two dissimilar data with a distance margin. With the same CNN architecture used in TSML-Hybrid, the implemented DrLIM in~\cite{jia2014caffe} projects the MNIST test data into the 2-dimensional and 3-dimensional spaces as in Fig.~\ref{fig5:drlimView}. DrLIM does not benefit from the manifold learning theory and thus provides only one perspective to view the outputs.

\textbf{Running time}: using the deep learning tool Caffe~\cite{jia2014caffe} in CPU mode only, we carried out all the experiments on a machine with a 2.3 GHz dual-core CPU, 4GB RAM and 64-bit operating system. In general, as DrLIM and TSML-Hybrid have the same deep CNN architecture, performing gradient descent once for either the triangular loss or the contrastive loss costs almost the same time. However, the hybrid training strategy helps TSML-Hybrid to train a model much faster. Concretely, the tiny-scale training costs about 150 seconds and the large-scale training costs about 960 seconds, hence obtaining the TSML-Hybrid models averagely costs about 1110 seconds (see Table~\ref{tal5:compTime}). For DrLIM that directly performs large-scale training on data pairs, the convergence speed slows down significantly than the other methods, the average training time for each DrLIM model is about 9600 seconds.


\textbf{Classification accuracy}: a comparison on classification performance is summarized in Table~\ref{tal5:compClass}. The classical CNN classifier -- the present CNN followed by a Euclidean loss layer or softmax loss layer -- is set as the baseline for classification. The proposed TSML-Hybrid method offers two views of the target space. Take a target space of dimension 10 for example, View 1 is the result of TSML-Hybrid with the output layer size equal to 10;
View 2 is the result of an output dimension 11 and then unfolded into the 10-dimensional space by coordinate transformation.

Firstly, we evaluate all the methods in the 10-dimensional space because the output of CNN is determined by the number of classes. Comparing these results we find that our TSML-Hybrid method obtains comparable classification results with the CNN classifier ($98.97\%$) in the 10-dimensional space ($99.02\%$ and $98.87\%$ in the two views respectively). However, DrLIM relatively performs worse ($98.35\%$) than the other three. This is because DrLIM trains on a limited number of data pairs and does not fully make use of all the labeled data samples.

Concerning the lower dimensional spaces, the CNN classifier failed for dimensionality reduction because the dimension of the handcrafted target vectors was fixed. DrLIM and TSML-Hybrid relaxes this constraint by metric learning and realizes classification in the visualizable spaces. Generally, TSML-Hybrid (View 2) gets better results than DrLIM and TSML-Hybrid (View 1) in these low dimensional spaces.
For DrLIM and TSML-Hybrid (View 1), the classification accuracy decreases as the target dimension is reduced and a large decline occurs when the target dimension changes from 3 to 2. By using coordinate transformation, TSML-Hybrid (View 2) delays the decline till the dimension reduced from 2 to 1. Therefore, the View 2 is preferred if the classification performance is mainly concerned.

\begin{table}[t]
\caption[Comparison on training time of different methods]{Comparison on training time (in seconds) of different methods. '---' means no available result. View 1: before unfolding; View 2: after unfolding. }
\centering
\smaller{
\begin{tabular}{cccccc}
\toprule
Methods  & 1-dimension & 2-dimension & 3-dimension & 10-dimension & Mean \\
\midrule
CNN classifier~\cite{jia2014caffe} & --- & --- & --- & 968.33 & 968.33\\
DrLIM~\cite{hadsell2006dimensionality}  &--- & 9632.10 & 9579.83 & 9556.59 & 9589.51 \\
TSML-Hybrid (view 1) & --- & 1111.64 & 1113.89 & 1119.19 & \multirow{2}*{1116.75} \\
TSML-Hybrid (view 2)  & 1111.64 & 1113.89 & 1121.10 & 1117.92 & ~ \\
\bottomrule
\end{tabular}}
\label{tal5:compTime}
\end{table}

\begin{table}[t]
\caption[Comparison on classification accuracy of different methods]{Comparison on classification accuracy of different methods. '---' means no available result. View 1: before unfolding; View 2: after unfolding. }
\centering
\begin{tabular}{ccccc}
\toprule
Methods  & 1-dimension & 2-dimension & 3-dimension & 10-dimension \\
\midrule
CNN classifier~\cite{jia2014caffe} & --- & --- & --- & 98.97\% \\
DrLIM~\cite{hadsell2006dimensionality}  &--- & 94.16\% & 97.28\% & 98.35\%  \\
TSML-Hybrid (view 1) & --- & 94.76\% & 98.05\% & \textbf{99.02\%}  \\
TSML-Hybrid (view 2)  &\textbf{94.76\%} & \textbf{98.05\%} & \textbf{98.66\%} & 98.87\% \\
\bottomrule
\end{tabular}
\label{tal5:compClass}
\end{table}



\vspace{-5pt}
\section{Conclusion}
\label{sec5:conclusion}
\vspace{-5pt}
In this paper, we introduced a novel Triangular Similarity to define the pairwise data relationship in the target spaces and deploy it into the CNN architecture to project an image to a point in low dimensional spaces. Interestingly, this similarity metric learning approach enables us to make use of manifold learning theories and visualize the output data in two different views.
We succeeded in mapping thousands of images to local parts of low dimensional spaces but having a competitive classification accuracy.
In the near future, we would like to apply the method on other datasets and extend our work to unsupervised tasks.

\bibliographystyle{splncs}
\bibliography{egbib}


\end{document}